\documentclass{article}

\usepackage{microtype}
\usepackage{graphicx}
\usepackage{subfigure}
\usepackage{booktabs}

\usepackage{hyperref}

\usepackage{microtype}
\usepackage{amsfonts}
\usepackage{graphicx}
\usepackage {amsmath}
\usepackage{booktabs}
\usepackage[accepted]{icml2023}

\usepackage{amsmath}
\usepackage{amssymb}
\usepackage{mathtools}
\usepackage{amsthm}

\usepackage{hyperref}
\usepackage{colortbl}
\usepackage{times,graphicx,amsmath,amssymb,booktabs,tabulary,multirow,overpic,xcolor}
\usepackage[capitalize,noabbrev]{cleveref}

\theoremstyle{plain}

\theoremstyle{definition}

\theoremstyle{remark}

\usepackage[textsize=tiny]{todonotes}

\icmltitlerunning{~ \hfill Unlocking Masked Autoencoders as Loss Function \hfill \thepage}

\begin{document}

\twocolumn[
\icmltitle{Unlocking Masked Autoencoders as Loss Function
\\ for Image and Video Restoration}



\icmlsetsymbol{equal}{*}

\begin{icmlauthorlist}
\icmlauthor{Man Zhou}{equal,1}
\icmlauthor{Naishan Zheng}{equal,2}
\icmlauthor{Jie Huang}{2}
\icmlauthor{Chunle Guo}{3}
\icmlauthor{Chongyi Li}{1}
\end{icmlauthorlist}

\icmlaffiliation{1}{S-Lab, Nanyang Technological University, Singapore} 
\icmlaffiliation{2}{University of Science and Technology of China, China} 
\icmlaffiliation{3}{Nankai University, China}

\icmlcorrespondingauthor{Chongyi Li}{chongyi.li@ntu.edu.sg}

\icmlkeywords{Machine Learning, ICML}

\vskip 0.3in
]



\printAffiliationsAndNotice{\icmlEqualContribution} 

\begin{abstract}
Image and video restoration has achieved a remarkable leap with the advent of deep learning. The success of deep learning paradigm lies in three key components: data, model, and loss. Currently, many efforts have been devoted to the first two while seldom study focuses on loss function. With the question ``are the de facto optimization functions e.g., $L_1$, $L_2$, and perceptual losses optimal?'',  we explore the potential of loss and raise our belief ``\textbf{learned loss function empowers the learning capability of neural networks for image and video restoration}''.

Concretely, we stand on the shoulders of the masked Autoencoders (MAE) and formulate it as a `learned loss function', owing to the fact the pre-trained MAE innately inherits the prior of image reasoning. We investigate the efficacy of our belief from three perspectives: 1) from task-customized MAE to native MAE, 2) from image task to video task, and 3) from transformer structure to convolution neural network structure.  Extensive experiments across multiple image and video tasks, including image denoising, image super-resolution, image enhancement, guided image super-resolution, video denoising, and video enhancement, demonstrate the consistent performance improvements introduced by the learned loss function. Besides, the learned loss function is preferable as it can be directly plugged into existing networks during training without involving computations in the inference stage. Code will be publicly available.

\end{abstract}

\begin{figure}[ht]
\centering
\includegraphics[width=\columnwidth]{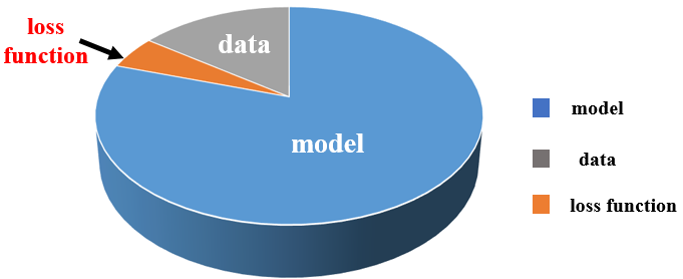}
\caption{Three key components, data, model, and loss, in the deep learning era. More efforts have been devoted to the data and model while seldom study focuses on the remaining loss function that is essential for optimizing deep models. Such observation motivates us to explore the potential of loss function.}
\label{abs}
\end{figure}

\begin{figure}[!t]
\centering
\includegraphics[width=0.85\linewidth]{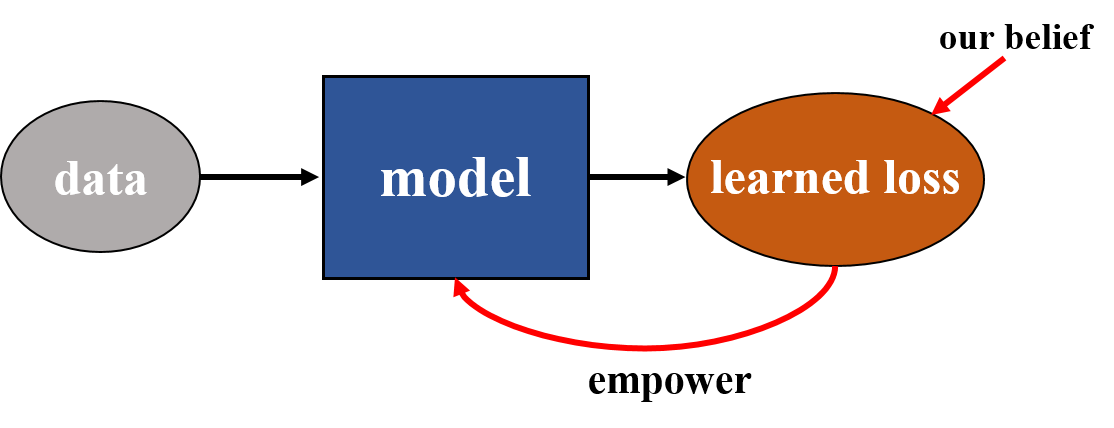}
\caption{Thumbnail of the proposed belief ``learned loss function empowers the model learning capability''.}
\label{fea}
\end{figure}

\section{Introduction}
\label{sec:intro}
Image and video restoration is a long-standing problem in computer vision. 
In the artificial intelligence era, deep learning has been instrumental in igniting the  image and video restoration boom \cite{ren2016single, li2018benchmarking, li2019underwater, yan2017image, ren2016image, pan2016blind, liu2021retinex, liu2018learning, wang2019edvr, liang2022vrt}.
As well recognized, deep learning paradigm involves three key components: \textbf{data}, \textbf{model}, and \textbf{loss}. 
Considerable efforts have been devoted to the study of data and model, e.g., collecting large-scale and high-quality datasets and designing different network structures. 
However, seldom study focuses on loss function, as illustrated in Figure \ref{abs}.
It seems that  $L_1$, $L_2$, and perceptual losses are the default optimization functions in supervised image and video restoration.
This begs the question, are these de facto optimization functions optimal?

The widely-used loss functions, e.g., $L_1$ and $L_2$, are born without `energy', i.e.,  they are only the measurement tools in purely mathematical form and are without the knowledge associated with  restoration task. Hence, we argue that they may be sub-optimal.
Previous approaches attempt to introduce `energy' into loss function. For example, perceptual loss \cite{johnson2016perceptual} employs the VGG network pre-trained on ImageNet as the perceptual measurement function. 
Since the VGG network can represent perceptual and semantic information via pre-training on the large-scale dataset, it intuitively has the capability of transferring such knowledge to downstream tasks when it is used as the loss function. 
However, the perceptual loss suffers from the perceptual-distortion trade-off \cite{johnson2016perceptual, liu2019classification}, i.e., improving the visual performance at the cost of the PSNR and SSIM scores. 
We therefore wonder ``whether we can provide a new solution to relieve such dilemma from the view of optimization function?''. 
In comparison to data collection and network designs, an elegant optimization function  is more appealing as it can be directly plugged into existing networks during training without involving computations in the inference stage.

Towards this goal,  we  explore the potential of loss function and present our belief ``learned loss function empowers the learning capability of neural networks for image and video restoration'', as shown in Figure \ref{fea}. 
As a simple and alternative solution in our belief,  we formulate the masked Autoencoders (MAE) as `learned loss function'. 
We find that the pre-trained MAE innately inherits the prior of image reasoning, benefiting for optimizing restoration models.
We believe the future of much more feasible solutions aligned with our belief. 

Specifically, we first investigate the efficacy of our belief from three perspectives: 1) from task-customized MAE to native MAE, 2) from image task to video task, and 3) from transformer structure to convolution neural network structure.
We further demonstrate the advantages of the learned loss via extensive experiments, including image de-noising, image super-resolution, image enhancement,  guided image super-resolution, video de-noising, and video enhancement. 
As shown in Figure \ref{fig:graphagg} and Figure \ref{fig:graphagg1}, we show a set of comparisons on the representative tasks of image super-resolution and image de-noising. As can be observed, equipped  with the learned loss function, the corresponding baselines are capable of achieving better visual awareness (especially eliminating the resulting gird effect of perceptual loss) for image super-resolution and removing the noise for image de-noising.

Our contributions are summarized as follows: \textbf{(1)} We propose a simple yet effective paradigm for image and video restoration, which is inspired by our belief ``learned loss function empowers the  learning capability of neural networks for image and video restoration''. \textbf{(2)} The proposed paradigm, especially the customized MAE, is capable of  improving the qualitative and  quantitative performance simultaneously, and alleviating the perceptual-distortion trade-off of commonly-used perceptual loss. \textbf{(3)} This is the first attempt to improve the performance of image and video models without changing model and data and achieve state-of-the-art performance across multiple image and video restoration tasks.

\section{Related work}
\label{sec:related}

\subsection{Loss function for image and video resotration}

In terms of image and video restoration tasks, the $\rm L_1$ and $\rm L_2$ have been widely employed as the loss function to measure the distance between the model output and the corresponding ground truth. However, they are born without energy, only the measuring tools in pure mathematical form. Moreover, they do not have the intrinsic knowledge associated with the optimization task. Since then, Justin \emph{et al} \cite{johnson2016perceptual} attempted to develop the pre-trained VGG network over image classification task as the perceptual measuring function. Owing to the VGG network that has already learned to encode the perceptual and semantic information on additional data samples, it naturally has the function to transfer the contained knowledge of the VGG network to the task model, \emph{e.g.,} image super-resolution task and style transformation task. However, the VGG loss suffers from the perceptual-distortion trade-off \cite{johnson2016perceptual,liu2019classification} where the corresponding visual results have been improved at the cost of the PSNR and SSIM performance degradation. In addition, the gap between the image-classification pre-trained VGG network and the regressive image/video restoration tasks inherently exists and thus disturbs the model optimization. This motivates us to explore the regressive learned loss function (acting as image prior) to solve the above issues.

\subsection{Mask image modeling}

To implement our belief in a regressive manner where the learned loss function can be treated as image prior, the first choice is the masked autoencoders (MAE) proposed by He \emph{et al} \cite{2022Masked}. MAE is a self-supervised learner of computer vision. The MAE masks random patches from the input image in a large ratio and reconstructs the missing patches in the pixel space. To highlight, MAE infers complex and holistic reconstructions, suggesting that it has learned numerous visual concepts and hypothesizing that this behavior occurs by rich hidden representations inside the MAE. Since then,  He \emph{et al} studied a conceptually simple extension of Masked Autoencoders (MAE) to spatiotemporal representation learning from videos \cite{feichtenhofer2022masked} by randomly masking out spacetime patches in videos and learning an autoencoder to reconstruct them in pixels. This study suggests that the general framework of masked autoencoding can be a unified methodology for representation learning with minimal domain knowledge. In this work, we stand on the shoulder of MAE to provide a simple and alternative solution to our belief. 

\begin{figure*}[t]
	\centering
	\includegraphics[width=\textwidth]{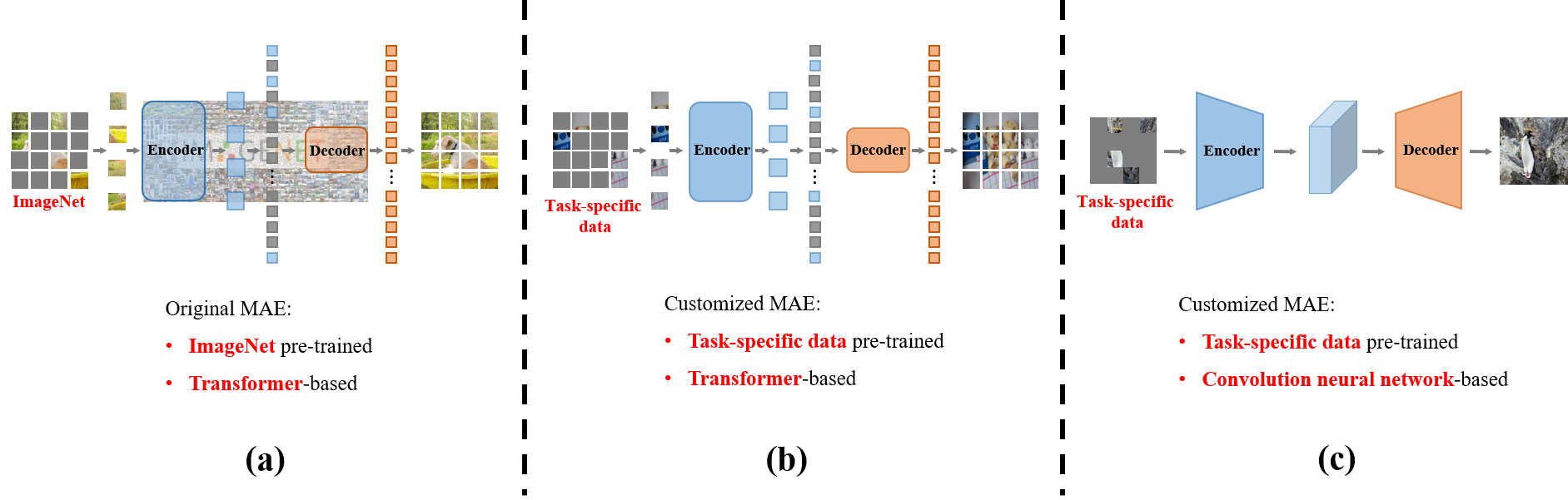}
	\caption{\textbf{The developed variants of MAE.} In detail, (a) depicts the original MAE proposed by He \emph{et al} \cite{2022Masked}, (b) represents the customized version of MAE pre-trained on task-specific data, and (c) is tailored by replacing the transformer architecture of the original MAE with the pure convolution neural network.}
	\label{mainfig}
\end{figure*}

\section{Methods}
In this section, we will first revisit the original image-version Masked Autoencoders (MAE) \cite{2022Masked} and the video extension of Masked Autoencoders \cite{feichtenhofer2022masked}, then detail the tailored MAE for the specific tasks. 

Our proposed belief is ``learned loss function empowers the modeling learning capability''. Therefore, the corresponding pipeline consists of two steps: 1) firstly train the additional network to learn the image prior, and 2) employ the pre-trained additional network as the loss function to guide the learning of task-specific networks. In this work, we treat MAE as the additional loss function network and conduct image and video restoration study.

\subsection{Single image restoration}
\noindent\textbf{Three configurations.} In terms of single image restoration tasks, we adopt the image-version MAE and then redevelop MAE as the ``learned loss function''. As shown in Figure \ref{mainfig}, the core designs are with respect to the following three levels: 

\begin{itemize}
  \item directly transform the encoder part of the original MAE pre-trained on the ImageNet dataset as the loss function in Figure \ref{mainfig} (a);

 \item train the original MAE on task-specific data and then employ the well-trained encoder part as the loss function in Figure \ref{mainfig} (b);
  
  \item replace the transformer architecture of the original MAE with the pure convolution neural network, follow the masking patch strategy for training and employ the encoder part of the well-trained one as the loss function in Figure \ref{mainfig} (c).  
\end{itemize}

\noindent\textbf{Implementation details.} The aforementioned first two configurations follow the same setting as the original MAE: 

\begin{itemize}
  \item adopt the asymmetric designs of transformer as the encoder and decoder;

 \item divide an image into regular non-overlapping patches, randomly sample a small subset of patches, and mask the remaining ones;
  
  \item the encoder is applied to the small subset of visible patches. Mask tokens are introduced after the encoder, and the full set of encoded patches and mask tokens is processed by a small decoder that reconstructs the original image in pixels; 
  
  \item the input of the encoder is the image patch, not the whole image architecture.
\end{itemize}

The remaining configuration employs the pure convolution neural network with masking patch strategy: 

\begin{itemize}
  \item adopt the pure convolution neural network as the encoder and decoder; 
  
 \item evenly divide the input image, randomly sample a small subset of regions, and mask the remaining ones while keeping the whole image architecture;
  
  \item  both the small subset of visible patches and mask tokens are  processed by the encoder and decoder that reconstructs the original image in pixels; 
  
  \item the input of the encoder is the whole image architecture, not the image patch.
\end{itemize}

\noindent\textbf{Hypothesis.} The redeveloped MAE is pre-trained on the ImageNet dataset or the task-specific ground truth in a self-supervised manner. Therefore, the pre-trained MAE learns the intrinsic knowledge (treated as the image prior) and then empowers the model learning capability by transferring the learned  intrinsic knowledge.

\begin{figure}[t]
	\centering
	\includegraphics[width=0.5\textwidth]{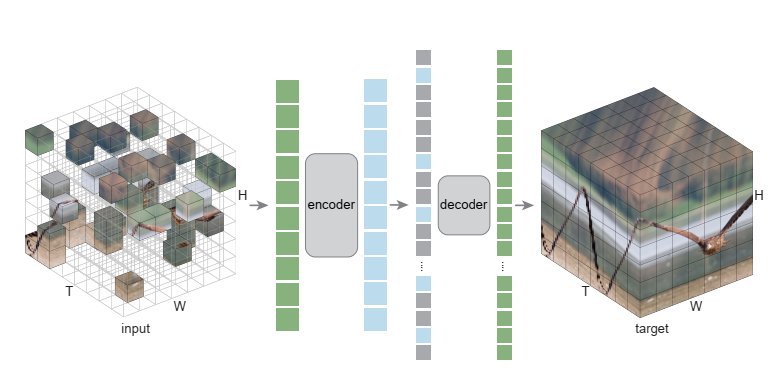}
	\caption{\textbf{Motivation.} Masked Autoencoders as spatiotemporal learners. We mask a large subset of random patches in spacetime. An encoder operates on a set of visible patches. A decoder then processes the full set of encoded patches and mask tokens to reconstruct the input. Except for the patch and positional embeddings, the encoder, decoder, and masking strategy have no  spatiotemporal inductive bias. }
	\label{videomae}
\end{figure}

\subsection{Video restoration}

Different from the single image restoration that only involves spatial information, video restoration tasks are required to explore the additional temporal information and achieve spatial-temporal consistency between the model outputs and the expected ones. Concerning our belief on single image restoration, the above image-version MAE is naturally extended to video-version MAE.

Similar to the aforementioned implementations, standing on the shoulders of the video-version extension of masked Autoencoders \cite{feichtenhofer2022masked}, we redevelop the masked image modeling as ``learned loss function'' to constrain the spatiotemporal representation consistency with respect to the following two levels:  

\begin{itemize}
  \item directly transform the encoder part of the original video-version MAE pre-trained on the Kinetics dataset as the loss function in Figure \ref{videomae};

 \item train the original MAE on task-specific video data and then employ the encoder part as the loss function in Figure \ref{videomae};
 
\end{itemize}

\noindent\textbf{Implementation details.} The aforementioned two configurations follow the same setting as the original video-version MAE: 

\begin{itemize}
  \item  randomly mask out spacetime patches in videos and then learn an autoencoder to reconstruct them;

  \item the only spacetime-specific inductive bias is on embedding the patches and their
positions; all other components are agnostic to the spacetime nature of the problem. In particular, the encoder and decoder are both vanilla Vision Transformers with no factorization or hierarchy, and our random mask sampling is agnostic to the spacetime structures.
  
\end{itemize}

\noindent\textbf{Hypothesis.} The redeveloped video-version MAE aims to reason the invisible patches from a small subset of visible patches in a self-supervised training manner. Therefore, the pre-trained MAE learns the intrinsic knowledge over both the spatial and temporal dimensions and then can be regarded as the loss function to  constrain the spatial-temporal consistency between the model output and ground truth, which is the core for video restoration tasks.

\subsection{Pipeline}

Suppose the task model as $f(\mathbf{x})$ that transforms the input image/video $\mathbf{x}$ to output  $\mathbf{y}$, the process can be written as
\begin{equation}
    \mathbf{y} = f(\mathbf{x}).
\end{equation}

Suppose the pre-trained MAE model as $f_{mae}(.)$ and its encoder part as $E_{mae}(.)$ that is employed as the complementary loss function to the original image-level loss function \emph{e.g.,} $\rm L_1$ and $\rm L_2$. The total loss function is remarked as 

\begin{equation}
    \mathbf{L} = ||\mathbf{GT}- \mathbf{y}||_{1,2} + \lambda ||E_{mae}(\mathbf{GT})-E_{mae}(\mathbf{y})||_{1,2},
\end{equation}
where $\lambda$ indicates the weighted factor and is set to $1$, $||.||_{1,2}$ is the image-level loss function \emph{i.e.,} $\rm L_1$ and $\rm L_2$, and $GT$ denotes the ground truth. To emphasize, $E_{mae}(.)$ represents the image-version MAE for image restoration tasks and the video-version MEA for  video restoration tasks, respectively.  

\begin{figure*}[ht]
	\centering
	\includegraphics[width=\textwidth]{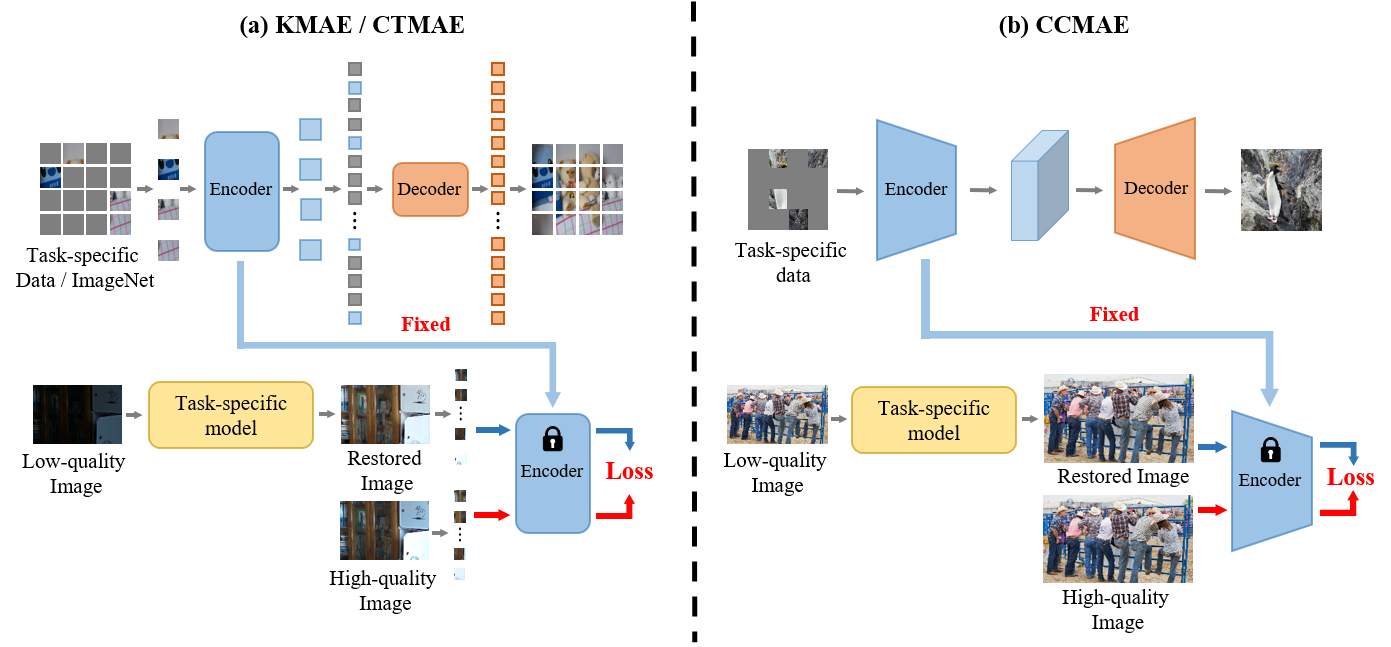} 
	\caption{The pipeline of the naive MAE loss implementation with the whole image input.}
	\label{w1}
\end{figure*}

\begin{figure}[t]
	\centering
	\includegraphics[width=\columnwidth]{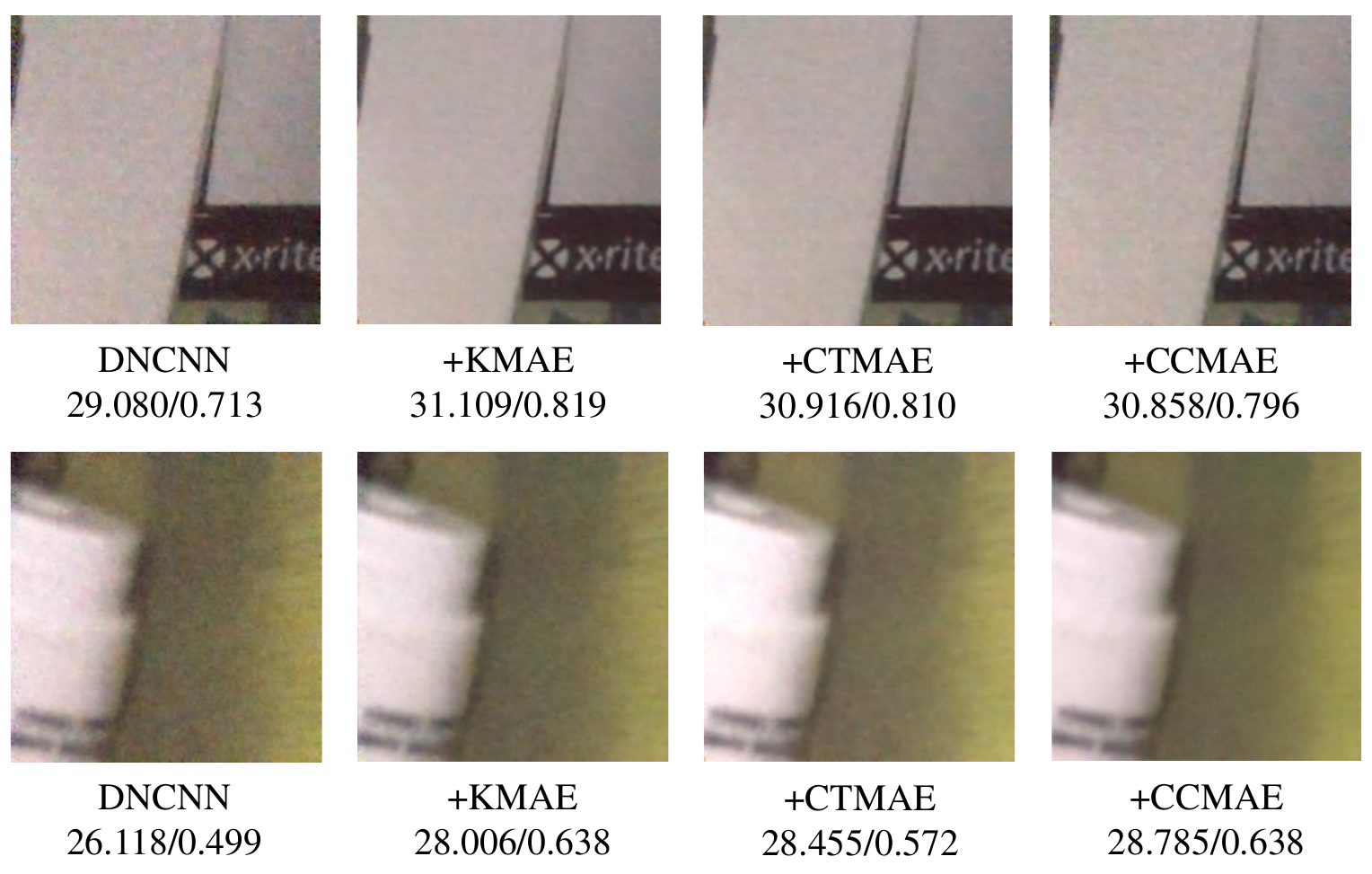}
	\caption{The visual comparison for the image de-noising. We also list the PSNR/SSIM scores under each case.}
	\label{fig:graphagg1}
 	
\end{figure}

\begin{table*}[!htb]
\small
\centering
\renewcommand{\arraystretch}{1.2}
\caption{\textbf{Quantitative comparisons of image enhancement.}}
\begin{tabular}{l l |c c c c c c }
    \hline
    \multirow{2}{*}{Model} & \multirow{2}{*}{Configurations}& \multicolumn{3}{c}{LoL} & \multicolumn{3}{c}{Huawei}  \\
    &&PSNR & SSIM & NIQE  & PSNR & SSIM  & NIQE\\
    \hline
    \multirow{5}{*}{SID} & Original& 20.2852 & 0.796 & 3.8667 & 16.9575  & 0.5958 & 3.6313  \\
     &+VGG Loss & 20.0959 & \textbf{0.8105} & 3.8778 & 16.7935 & \textbf{0.5996} & 3.6797   \\ \hline
    &+KMAE Loss & 20.4641 & 0.7955 & 3.8735 & 17.0227 & 0.5949 & 3.5958 \\
    &+CTMAE Loss & 20.5163 & 0.7967 & 3.8664  & 17.0716 & 0.5953 & 3.5954 \\
    &+CCMAE Loss & \textbf{20.5337} & 0.7971 & \textbf{3.8658}  & \textbf{17.1275} & 0.5955 & \textbf{3.5950} \\
    \hline
    \multirow{5}{*}{DRBN} & Original& 19.8509
 & 0.7799 & 4.7738  & 16.2231  & 0.5989 & 4.3273 \\
     &+VGG Loss & 19.7877 & \textbf{0.8024} & 4.7779 & 16.1322  & 0.5991 & 4.5515\\ \hline
    &+KMAE Loss & 19.9235 & 0.7779 & 4.5927 & 16.8194 & \textbf{0.5997} & 4.1815 \\
    &+CTMAE Loss & 20.0151 & 0.7751 & 4.7285  & 16.8112  & 0.594 & 4.3134 \\
    &+CCMAE Loss & \textbf{20.1336} & 0.7758 & \textbf{4.7276}  & \textbf{16.8376}  & 0.5952 & \textbf{4.3127} \\
    \hline
\end{tabular}
\label{image enhancement}
\end{table*}

\begin{table*}[!t]
\begin{minipage}[c]{0.48\textwidth}
\small
\centering
\setlength\tabcolsep{3pt}
\renewcommand{\arraystretch}{1.4}
\caption{\textbf{Quantitative comparisons for image de-noising.}}
\begin{tabular}{l l |c c  c c  }
    \hline
    \multirow{2}{*}{Model} & \multirow{2}{*}{Configurations}& \multicolumn{2}{c}{DND} & \multicolumn{2}{c}{SIDD}  \\
    &&PSNR$\uparrow$ & SSIM$\uparrow$   & PSNR$\uparrow$ & SSIM$\uparrow$  \\
    \hline
    \multirow{5}{*}{DnCNN} & Original& 32.4249 & 0.8413   & 23.6612& 0.583  \\
     &+KMAE Loss & 33.9195 & 0.8756 &   25.5441    &  0.682 \\
    &+CTMAE Loss & 34.1275 & 0.8847 & 25.7317 &  0.723 \\
    &+CCMAE Loss & \textbf{34.2592} & \textbf{0.8861} & \textbf{25.8429}  & \textbf{0.741}   \\
    \hline
    \multirow{5}{*}{MPRnet} & Original& 39.2401
 & 0.9555 &   38.8405 & 0.951  \\
     &+KMAE Loss & 39.2586 & 0.9556 & 38.8617   & 0.953   \\
    &+CTMAE Loss & 39.2874 & 0.9561 & 38.9033 &  0.953\\
    &+CCMAE Loss & \textbf{39.2983} & \textbf{0.9578} & \textbf{38.9184}   & \textbf{0.955}   \\
    \hline
\end{tabular}
\label{image de-noising}
\end{minipage}
\hspace{0.0001\linewidth}
\begin{minipage}[c]{0.48\textwidth}
\small
\centering
\setlength\tabcolsep{3pt}
\renewcommand{\arraystretch}{1.12}
\caption{\textbf{Quantitative comparisons for image super-resolution.}}
\begin{tabular}{l l |c c  c c  }
    \hline
    \multirow{2}{*}{Model} & \multirow{2}{*}{Configurations}& \multicolumn{2}{c}{$\times 2$} & \multicolumn{2}{c}{$\times 4$}  \\
    &&PSNR$\uparrow$  & NIQE$\downarrow$  & PSNR$\uparrow$   & NIQE$\downarrow$\\
    \hline
    \multirow{5}{*}{RFDN}  & Original & \textbf{34.1229}    & 4.2712  & \textbf{28.4997}
   & 6.3058  \\
     &+VGG Loss &  33.6880   & 4.4221 & 28.1066  & 6.6408  \\ \hline
    &+KMAE Loss &  34.0215  & 4.2307  & 28.3166  & 6.1563  \\
    &+CTMAE Loss & 33.9828   &  \textbf{4.1836} & 28.3018  & \textbf{5.1669}  \\
    &+CCMAE Loss & 34.0475    & 4.2254  & 28.2998  & 5.1827   \\
    \hline
       \multirow{5}{*}{EDSR} & Original & \textbf{34.4167}  & 4.1502 & \textbf{28.7324}   & 5.9066  \\
     &+VGG Loss &  34.1299    & 4.2712 & 28.3882  & 6.6583    \\ \hline
    &+KMAE Loss &  34.2711   & 4.1112 & 28.5879  & 5.7889  \\
    &+CTMAE Loss &   34.2597  & \textbf{4.0070} & 28.5475  & \textbf{4.9887}  \\
    &+CCMAE Loss &  34.3269    & 4.1431 & 28.5336  & 5.0011  \\
    \hline
\end{tabular}
\label{image super-resolution}
\end{minipage}
\end{table*}

\begin{table*}[t]
\small
\centering
\renewcommand{\arraystretch}{1.221}
\caption{\textbf{Quantitative comparisons of pan-sharpening.}}
\resizebox{0.96\linewidth}{!}{ 
\begin{tabular}{l l |c c c c  c c c c}
    \hline
    \multirow{2}{*}{Model} & \multirow{2}{*}{Configurations}& \multicolumn{4}{c}{WorldView-II} & \multicolumn{4}{c}{GaoFen2} \\
    & &PSNR$\uparrow$ & SSIM$\uparrow$&SAM$\downarrow$&ERGAS$\downarrow$ &PSNR$\uparrow$ & SSIM$\uparrow$&SAM$\downarrow$  &EGAS$\downarrow$ \\
    \hline
    \multirow{3}{*}{INNformer} & Original& 41.6903 & 0.9704 & 0.0227 & 0.9514         & 47.3528 & 0.9893 & 0.0102 & 0.5479  \\
                    
    &  +CTMAE Loss & \textbf{41.8559} & 0.9716 & 0.0223 & \textbf{0.9276}  & 47.4272 & 0.9901 & 0.0101 & 0.5356   \\
    
    &+CCMAE Loss & 41.8316 &\textbf{0.9717} & \textbf{0.0222} & 0.9276      & \textbf{47.4661} & \textbf{0.9903} & \textbf{0.0100} & \textbf{0.5356}  \\
    \hline
    
    \multirow{3}{*}{SFINet} &  Original& 41.7244 & 0.9725 & 0.0220 & 0.9506       & 47.4712 & 0.9901 & 0.0102 & 0.5462 \\
                            
    & +CTMAE Loss& \textbf{41.9514} & \textbf{0.9727} & 0.0218 & 0.9276     & 47.6227 & 0.9914 & 0.0101 & 0.5277  \\
   
    & +CCMAE Loss& 41.9493 & 0.9726 & \textbf{0.0218} & \textbf{0.9276}    & \textbf{47.6318} & \textbf{0.9917} & \textbf{0.0100} & \textbf{0.5277}  \\
    \hline
\end{tabular}}
\label{tab:ps}
\end{table*}

\begin{table*}[!t]
\begin{minipage}[c]{0.48\textwidth}
\small
\centering
\setlength\tabcolsep{3pt}
\renewcommand{\arraystretch}{1.212}
\caption{\textbf{Quantitative comparisons for video de-noising.}}
\begin{tabular}{l l |c c  c c }
    \hline
    \multirow{2}{*}{Model} & \multirow{2}{*}{Configurations}& \multicolumn{2}{c}{DAVIS} & \multicolumn{2}{c}{Set8}  \\
    &&PSNR$\uparrow$ & SSIM$\uparrow$   & PSNR$\uparrow$ & SSIM$\uparrow$  \\
    \hline
    \multirow{3}{*}{FastDVDnet} & Original& 31.64
 & 0.9188 & 32.31  & 0.8917  \\
    &+KMAE Loss & 32.82 & 0.919 & 32.42  & 0.893  \\
    &+CTMAE Loss & \textbf{33.37} & \textbf{0.919} & \textbf{32.47}  & \textbf{0.893}  \\
    \hline
    
      \multirow{3}{*}{PaCNet} & Original& 35.62 & 0.9221 & 32.68 & 0.895 \\
    &+KMAE Loss & 35.75 & 0.935 & 32.81 & 0.898  \\
    &+CTMAE Loss & \textbf{35.83} & \textbf{0.935} & \textbf{32.83} & \textbf{0.901}  \\
    \hline
\end{tabular}
\label{video de-noising}
\end{minipage}
\hspace{0.0001\linewidth}
\begin{minipage}[c]{0.48\textwidth}
\small
\centering
\setlength\tabcolsep{3pt}
\renewcommand{\arraystretch}{1.212}
\caption{\textbf{Quantitative comparisons for video enhancement.}}
\begin{tabular}{l l |c c c c c c }
    \hline
    \multirow{2}{*}{Model} & \multirow{2}{*}{Configurations}& \multicolumn{2}{c}{e-VDS} & \multicolumn{2}{c}{DRV}  \\
    &&PSNR$\uparrow$ & SSIM$\uparrow$   & PSNR$\uparrow$ & SSIM$\uparrow$  \\
    \hline
    \multirow{3}{*}{MBLLVEN} & Original& 24.98
 & 0.83 & 26.88 & 0.802 \\
    &+KMAE Loss & 25.13 & 0.83 & 27.57 & 0.807  \\
    &+CTMAE Loss & \textbf{25.15} & \textbf{0.85} & \textbf{27.83} & \textbf{0.811}  \\
    \hline
    \multirow{3}{*}{SMOID} & Original& 25.17 & 0.85 & 24.05 & 0.782  \\
    &+KMAE Loss & 25.29 & 0.87 & 25.13 & 0.788  \\
    &+CTMAE Loss & \textbf{25.35} & \textbf{0.87} & \textbf{25.22} & \textbf{0.801}  \\
    \hline
\end{tabular}
\label{video enhancement}
\end{minipage}
\end{table*}

\section{Experiments}
To demonstrate the efficacy of our proposed belief ``learned loss function empowers the model learning capability", we conduct extensive experiments across multiple image and video tasks, including image super-resolution, image enhancement, image de-noising, guided image super-resolution, video enhancement, and video de-noising. The more results refer to \textbf{\textit{Appendix.}}

\subsection{Experimental settings} 
\label{4.1section}

\textbf{Single image super-resolution.} Following \cite{lim2017enhanced,Liu_2020_CVPR}, the widely-used high-quality DIV2K dataset is used for image super-resolution task.  DIV2K dataset consists of 800 training images with 2K resolution and covers diverse contents from cityscapes to natural scenarios.  In the experimental setting, we perform image super-resolution with the upscaling factors $2$ and $4$ and employ the representative EDSR \cite{lim2017enhanced} and RFDN \cite{10.1007/978-3-030-67070-2_2} as the baselines.

\textbf{Image enhancement.} We verify our belief on two popular image enhancement benchmarks, including LOL \cite{Chen2018Retinex} and Huawei \cite{hai2021r2rnet}. LOL dataset consists of 500 low-/normal-light image pairs, and we split 485 for training and 15 for testing.  Huawei dataset contains 2480 paired images, and we split 2,200 for training and 280 for testing. Further, we adopt the two promising image enhancement algorithms, SID \cite{chen2018learning} and DRBN \cite{yang2020fidelity} as the  baselines.

\textbf{Image De-noising.}  Following \cite{9577298}, to evaluate our belief on the image de-noising task, we employ the widely-used SIDD dataset as training benchmark. Further, the corresponding performance evaluation is conducted on the remaining validation samples from the SIDD dataset \cite{abdelhamed2018high} and the DND benchmark dataset \cite{plotz2017benchmarking}. Two representative image de-noising algorithms DnCNN \cite{zhang2017beyond} and MPRnet \cite{9577298} are selected as the baselines.

\textbf{Guided Image Super-resolution.} Following \cite{9662053,sfinet}, we adopt the pan-sharpening, the representative task of guided image super-resolution, for evaluations. The WorldView II, WorldView III, and GaoFen2 datasets  \cite{9662053,sfinet} are used for experimental implementations. We employ  two state-of-the-art methods as the baselines to validate the effectiveness of our belief, including  INNformer \cite{9662053} and SFINet \cite{sfinet}.

\textbf{Video image de-noising.} Following \cite{vaksman2021patch, fastdvdnet}, the popular video de-nosing datasets, DAVIS and Set8, are used as evaluation benchmarks. The DAVIS dataset \cite{arias2015towards} consists of 90 training video sequences and 30 testing ones while Set8 dataset \cite{fastdvdnet}. We employ the representative PaCNet \cite{vaksman2021patch} and FastDVDnet \cite{fastdvdnet} as the baselines.

\textbf{Video enhancement.} For this task, the corresponding e-VDS dataset \cite{chan2022generalization} and DRV dataset \cite{9010274} are employed for experimental evaluations. We employ the MBLLVEN \cite{chan2022generalization} and  SMOID \cite{9010274} as the baseline.

Several widely-used image quality assessment (IQA) metrics are employed to evaluate the performance, including the relative dimensionless global error in synthesis (ERGAS) \cite{ergas},  peak signal-to-noise ratio (PSNR),  spectral angle mapper (SAM) \cite{sam}, structural similarity (SSIM), and natural image quality evaluator (NIQE) \cite{mittal2012making}.

\subsection{Implementation details} \label{4.2section}

\textbf{Image super-resolution and image enhancement.} In terms of perceptual-aware image restoration tasks, \emph{e.g.,} image super-resolution and image enhancement, the implementation variants of the corresponding baselines are organized as the five configurations:
\begin{itemize}
\setlength{\itemsep}{0pt}
\setlength{\parsep}{0pt}
\setlength{\parskip}{0pt}

\item [1)] \textbf{Original}: the baseline with the original image-level loss function ($\rm L_1$ or $\rm L_2$); 

\item [2)] \textbf{+VGG Loss}: complementing the original image-level loss function  ($\rm L_1$ or $\rm L_2$)  with the VGG Loss, forming the total loss;

\item[3)] \textbf{+KMAE Loss}: complementing the original image-level loss function  ($\rm L_1$ or $\rm L_2$)  with the directly transformed MAE encoder as the total loss; 

\item[4)] \textbf{+CTMAE Loss}:  complementing the original image-level loss function  ($\rm L_1$ or $\rm L_2$)  with the well-trained transformer-based MAE encoder on task-specific data as the total loss;

\item[5)] \textbf{+CCMAE Loss}:   complementing the original image-level loss function  ($\rm L_1$ or $\rm L_2$)  with the well-trained convolution neural network-based MAE encoder on task-specific data as the total loss.

\end{itemize}

\textbf{Image de-noising.} Following the above settings, the implementation configurations on image de-noising are organized as the four variants:
\begin{itemize}
\setlength{\itemsep}{0pt}
\setlength{\parsep}{0pt}
\setlength{\parskip}{0pt}

\item [1)] \textbf{Original}

\item [2)] \textbf{+KMAE Loss}

\item[3)] \textbf{+CTMAE Loss}

\item[4)] \textbf{+CCMAE Loss}

\end{itemize}

\textbf{Guided image super-resolution.} Pan-sharpening, the representative task of guided image super-resolution in remote sensing field, is adopted for evaluation. As well recognized, the data manifold used on the original MAE is different from that of remote sensing data. Therefore, the pre-trained MAE on the natural manifold cannot be used as the loss constraint for pan-sharpening task. 

Based on the above analysis, the implementation variants of the corresponding baselines for pan-sharpening are organized as the three variants:
\begin{itemize}
\setlength{\itemsep}{0pt}
\setlength{\parsep}{0pt}
\setlength{\parskip}{0pt}

\item [1)] \textbf{Original}

\item[2)] \textbf{+CTMAE Loss}

\item[3)] \textbf{+CCMAE Loss}

\end{itemize}

\begin{table*}[t]
\small
\centering
\caption{\textbf{Extension of patch-version design for pan-sharpening.}}
\resizebox{0.72\textwidth}{!}{ 
\begin{tabular}{l l |c c c c}
    \hline
    \multirow{2}{*}{Model} & \multirow{2}{*}{Configurations}& \multicolumn{4}{c}{WorldView-II}  \\
    & &PSNR$\uparrow$ & SSIM$\uparrow$&SAM$\downarrow$&ERGAS$\downarrow$  \\
    
    \hline
    
    \multirow{3}{*}{SFINet} &  Original& 41.7244 & 0.9725 & 0.0220 & 0.9506        \\
                            
    & $\rm +P\_CTMAE$ Loss& 41.9371 & 0.9726 & 0.0218 & 0.9269       \\
   
    & $\rm +P\_CCMAE$ Loss& \textbf{41.9554} & \textbf{0.9727} & \textbf{0.0218} & \textbf{0.9252}      \\
    \hline
\end{tabular}}
\label{tab-patch}
\vspace{-0.6em}
\end{table*}

\textbf{Video de-noising and Video enhancement.} In terms of video tasks where  spatial-temporal consistency is the core ideology, the video-version MAE-based implementation variants of the corresponding baselines are organized as the three variants:
\begin{itemize}
\setlength{\itemsep}{0pt}
\setlength{\parsep}{0pt}
\setlength{\parskip}{0pt}

\item [1)] \textbf{Original}: the baseline with the original image-level loss function ($\rm L_1$ or $\rm L_2$) on the output sequence and the expected ones; 

\item[2)] \textbf{+KMAE Loss}: complementing the original image-level loss function ($\rm L_1$ or $\rm L_2$)   with the directly transformed video-version MAE encoder as the total loss; 

\item[3)] \textbf{+CTMAE Loss}:  complementing the original image-level loss function ($\rm L_1$ or $\rm L_2$)  with the well-trained transformer-based video-version MAE encoder on task-specific data as the total loss.

\end{itemize}

\vspace{0.2em}
\subsection{Comparison and analysis} \label{integration}

\textbf{Image super-resolution.}  The quantitative comparisons over image super-resolution are presented in Table \ref{image super-resolution}. As can be seen,  all the reported baselines integrated with our proposed MAE loss have achieved better  perceptual-distortion trade-off  across all the datasets, suggesting the effectiveness of our belief. In comparison to the VGG loss, all the MAE variants have obtained performance gains in terms of PSNR and perceptual-aware NIQE simultaneously. In terms of the baselines, all the MAE loss variants have achieved better NIQE at cost of a slight decline in PSNR. 

Figure \ref{fig:graphagg} illustrates the visual results. Compared with the VGG loss, our proposed MAE loss is capable of 
removing the grid effect which is commonly introduced by  VGG loss, thus achieving better perceptual awareness. The reported NIQE also demonstrates these visual clues and validates that MAE loss alleviates the perceptual-distortion trade-off caused by the VGG loss.

 \vspace{0.1em}
 \textbf{Guided image super-resolution.}  The quantitative results for pan-sharpening are summarized in Tables \ref{tab:ps} where the best results are highlighted in bold. From the results, by integrating with our proposed MAE variants, all the reported baselines have  achieved consistent performance gains across all the datasets in terms of  all metrics, suggesting the effectiveness of our belief. 

\vspace{0.1em}
 \textbf{Image enhancement and de-noising.}  The quantitative results are presented in Tables  \ref{image enhancement} and \ref{image de-noising}.  From the results, by integrating with our proposed belief, we can observe the consistent performance gain against the baselines across all the datasets in image enhancement and de-noising tasks, suggesting the effectiveness of our belief.  Specifically, in terms of image enhancement, the baseline DRBN with the VGG loss has obtained the degradation for both PSNR  and perceptual-aware NIQE. Instead, the DRBN integrated with our proposed variants of MAE loss has achieved performance gains over the above metrics. It validates that the MAE loss is capable of improving the qualitative and quantitative performance simultaneously, alleviating the perceptual-distortion trade-off of the commonly-used perceptual loss. 

Observing two representative examples in Figure \ref{fig:graphagg1}, we can see that the results of the baseline DNCNN integrated with our belief   are cleaner than these of the original baseline and are with fewer noisy effects.  The
quantitative comparisons from the reported PSNR and SSIM also verify the above analysis.

 \vspace{0.1em}
 \textbf{Video enhancement and de-noising.} We perform the model performance comparison over different configurations, as described in the implementation details. The quantitative results are presented in Tables \ref{video de-noising} and \ref{video enhancement} where the best results are highlighted in bold. From the results, by integrating with our proposed video-version MAE variants, we can observe the consistent performance gain against all the baselines across all the datasets in all tested video tasks. The results suggest the effectiveness of our belief. It is attributed to the positive effect of the video-version MAE on maintaining the spatiotemporal consistency that is key for video restoration tasks.    

\subsection{Extension of patch-version designs}

In addition to feeding the whole model output and ground truth into the pre-trained MAE for measurement, the alternative strategy is to randomly select the image patch of model output and ground truth and feed them into the pre-trained MAE. It can be treated as ``regressive-version patch generative adversarial networks''. 

In our work, we only experiment with the above patch strategy over the pan-sharpening task by randomly selecting partial region to enable consistency. As shown in Table \ref{tab-patch} where $\rm +P\_CTMAE$ and $\rm +P\_CCMAE$ denote the corresponding customized MAE $\rm +CTMAE$ and MAE $\rm +CCMAE$ with patch sampling strategy,  the experimental results also demonstrate the consistent performance gain by introducing the patch-version designs.

\vspace{0.12em}
\section{Conclusion}
In this paper, orthogonal to the existing data and model studies, we instead resort our efforts to investigate the potential of loss function and present our belief ``learned loss function empowers the learning capability of neural networks''. Based on our belief, we provide a simple and alternative solution as an example. Specifically, we stand on the shoulders of the masked Autoencoders and redevelop the masked image modeling as ``learned loss function''.  Our proposed belief can be directly plugged into existing image and video restoration networks. Extensive experiments across multiple image and video restoration tasks demonstrate the consistent performance gains obtained by introducing our belief. 

\vspace{0.2211em}
We are optimistic about the future of our advocated belief. 
We plan to extend the proposed belief to problems involving natural language processing and to investigate its underlying  mechanisms to efficiently handle large-scale and multi-modality inputs such as images, audio, and video. In addition, we believe that our belief will hit the efficient learning field  with significant performance improvement, instead of violently stacking network architectures. 
Finally, we wish to promote the development of loss function and have to reiterate our belief ``learned loss function empowers the model learning capability''.

\nocite{langley00}

\bibliography{example_paper}
\bibliographystyle{icml2023}

\newpage
\appendix
\onecolumn
\section{\textbf{Appendix}.}
This supplementary document is organized as follows:

Sec. 1 provides the detailed implementation of our pipeline about how to integrate the pre-trained MAE into existing methods.

Sec. 2 provides the sufficient visual comparison over the representative tasks.

Sec. 3 discusses the perceptual-distortion trade-off, resulted from the perpetual-aware VGG loss.

Sec. 4 discusses the extension of patch-version designs.

Sec .5 presents the broader impacts.

\section{Technique pipeline.}
\label{discussion}

\textcolor{red}{The pipeline of the naive MAE loss implementation with the whole image input.} It consists of two steps: 1) pre-training the MAE to learn the intrinsic knowledge and 2) employ the encoder part of the well-trained MAE as the measurement tooling within the loss function, acting as the regularization term. To highlight, the model output and ground truth are fed into MAE encoder. As shown in Figure \ref{w1}, the left transformer-based encoder receives all the patch generated from the whole model output and ground truth. In the right part, the encoder is implemented by convolution neural networks and thus receives the whole image as input for measurement.

\begin{figure*}[h]
	\centering
	\includegraphics[width=\textwidth]{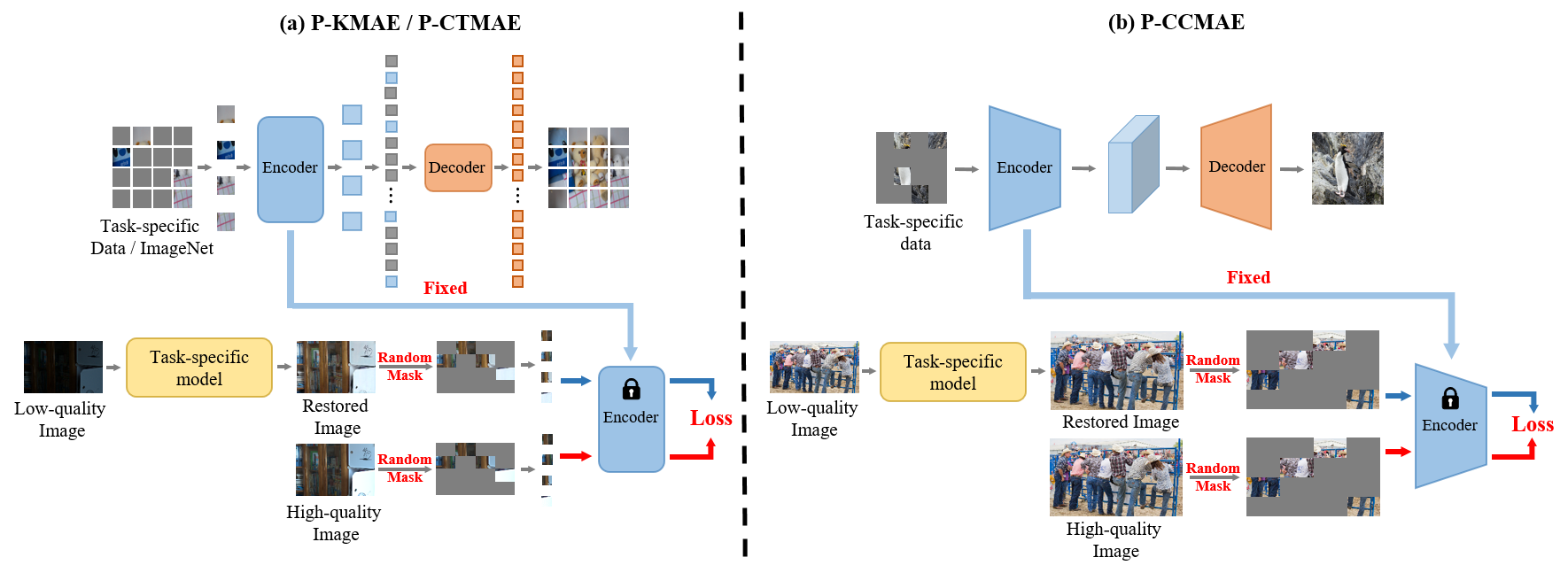} 
	\caption{The pipeline of patch-version MAE loss implementation with the image patch  input.}
	\label{p1}
\end{figure*}

\textcolor{red}{The pipeline of the patch-version MAE loss implementation with the image patch input.} In addition to feeding the whole model output and ground
truth into the pre-trained MAE for measurement, the alterna-
tive strategy is to randomly select the image patch of model
output and ground truth and feed them into the pre-trained
MAE. It can be treated as “regressive-version patch genera-
tive adversarial networks”.  It consists of two steps: 1) pre-training the MAE to learn the intrinsic knowledge and 2) employ the encoder part of the well-trained MAE as the measurement tooling within the loss function, acting as the regularization term. To highlight, the model output and ground truth are fed into MAE encoder. As shown in Figure \ref{p1}, the left transformer-based encoder receives the partially selected patches generated from the whole model output and ground truth. In the right part, the encoder is implemented by convolution neural networks and thus receives the masked image as input for measurement.

\begin{figure*}[h]
	\centering
	\includegraphics[width=\textwidth]{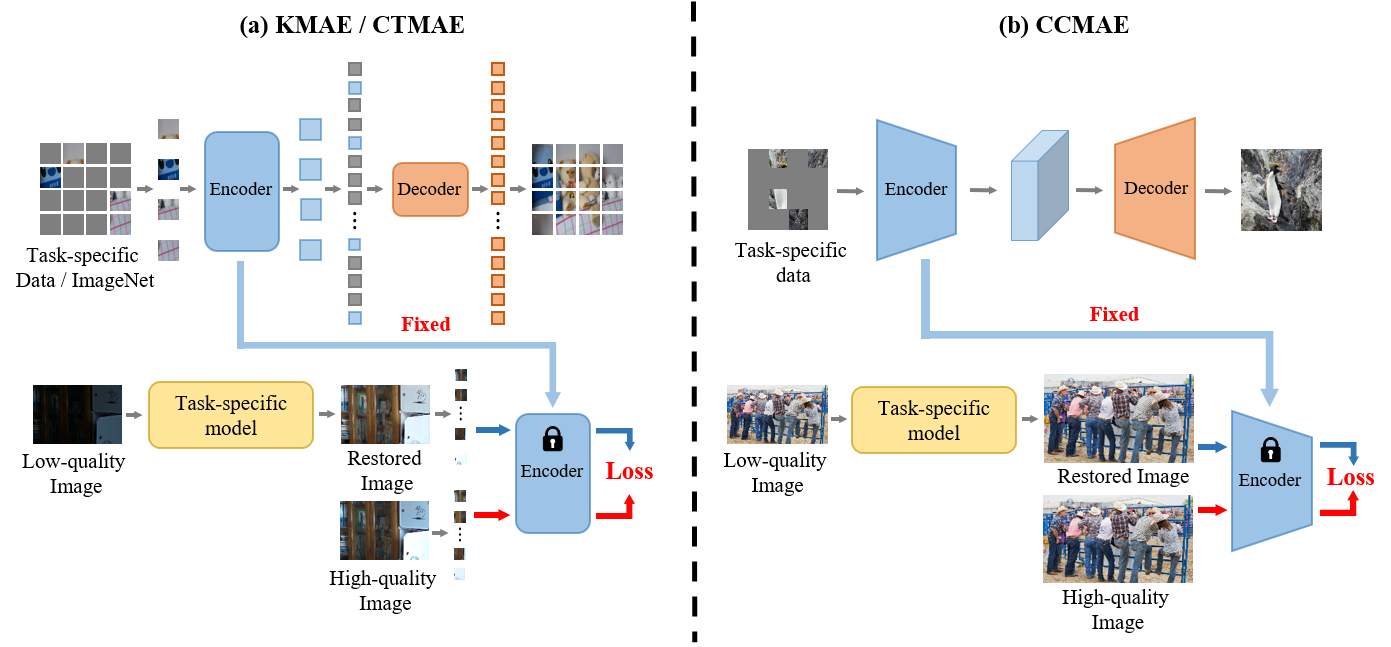} 
	\caption{The pipeline of the naive MAE loss implementation with the whole image input.}
	\label{w1}
\end{figure*}

\begin{figure*}[t]
	\centering
	\includegraphics[width=\textwidth]{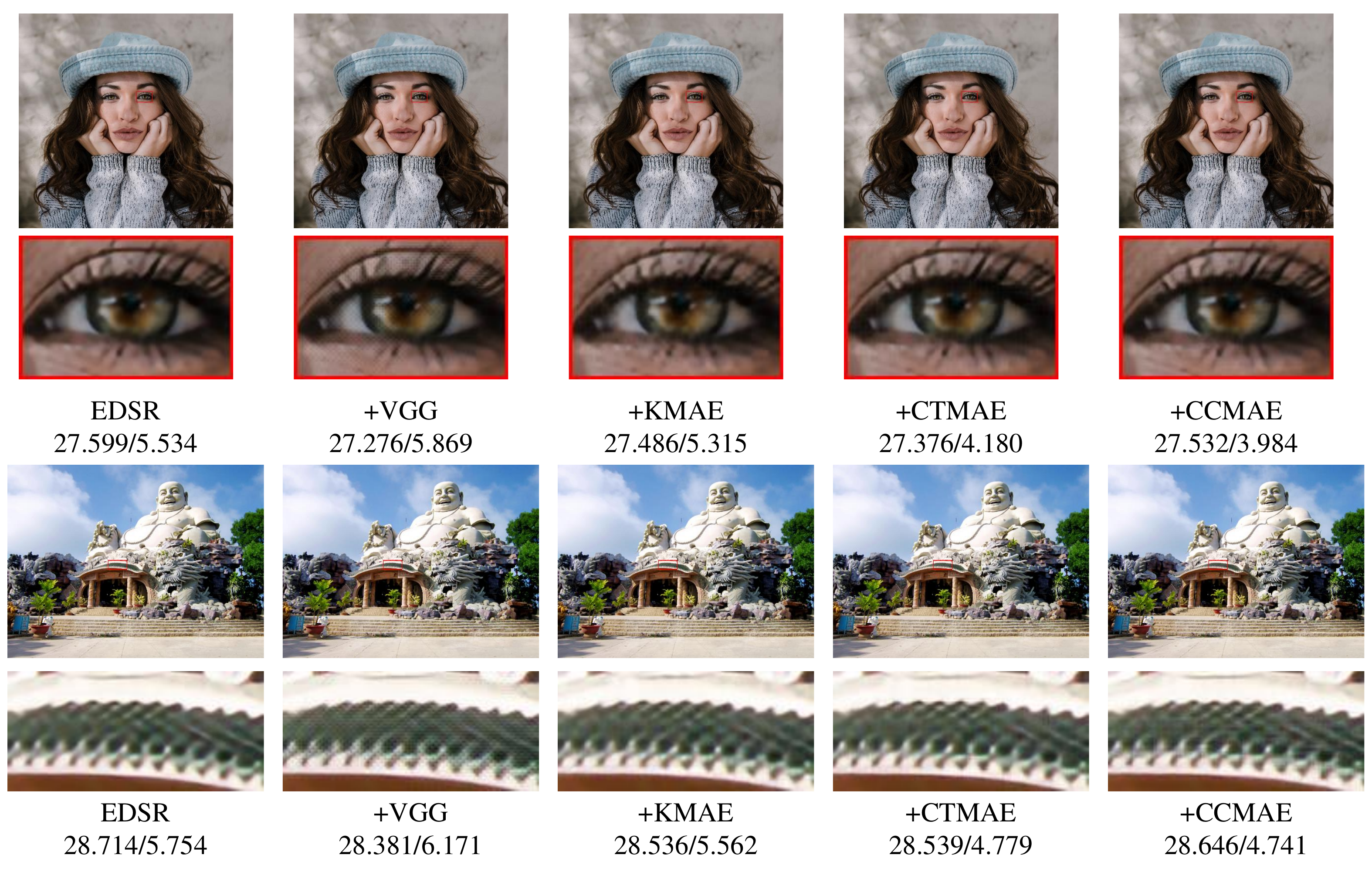}
	\caption{The visual comparison on $\times 4$ times image super-resolution. We also show the PSNR/NIQE scores under each case.  Zoom in for best view.}
	\label{fig:graphagg}

\end{figure*}

\section{Visual comparison.}
\label{illuPrior}
\begin{figure}[t]
	\centering
	\includegraphics[width=\columnwidth]{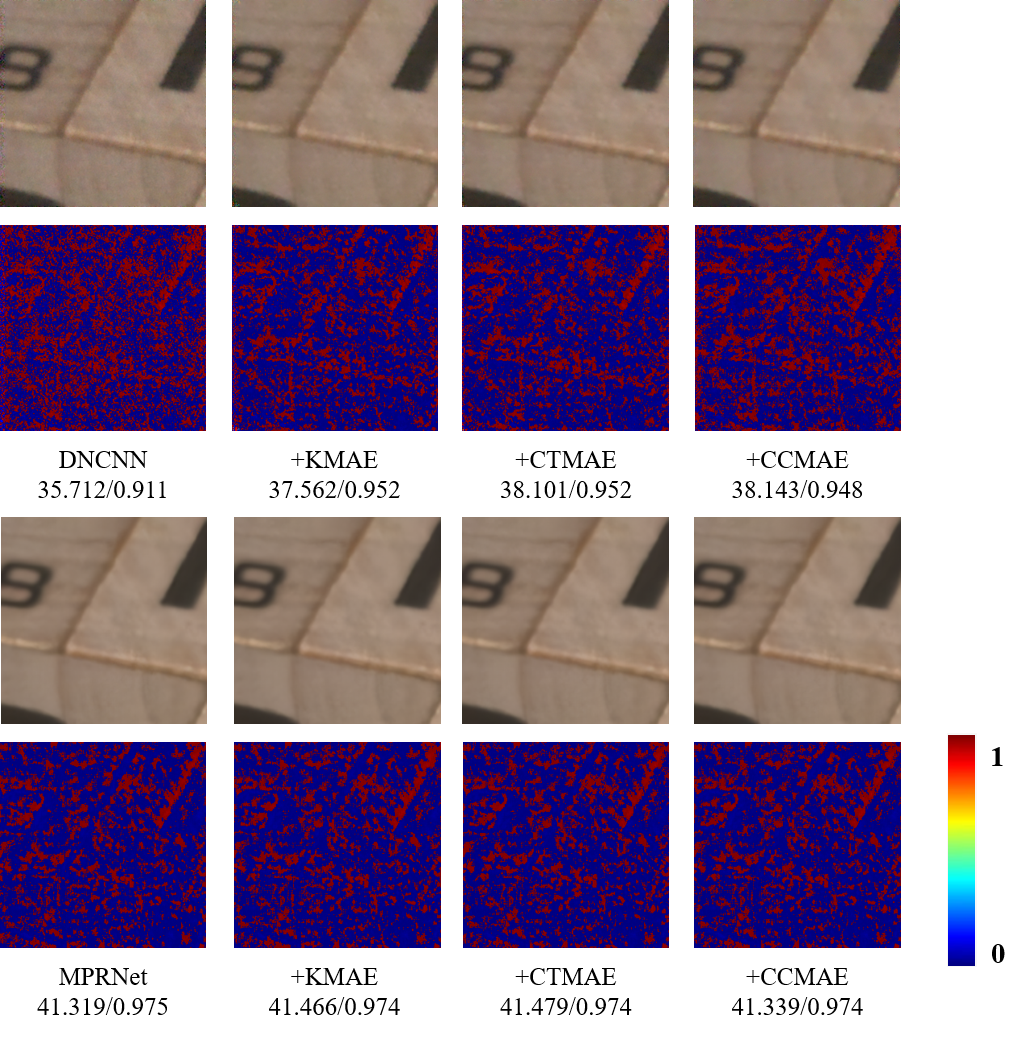}
	\caption{The visual comparison for the image de-noising. We also list the PSNR/SSIM scores under each case.}
	\label{dn1}
\end{figure}

\begin{figure}[t]
	\centering
	\includegraphics[width=\columnwidth]{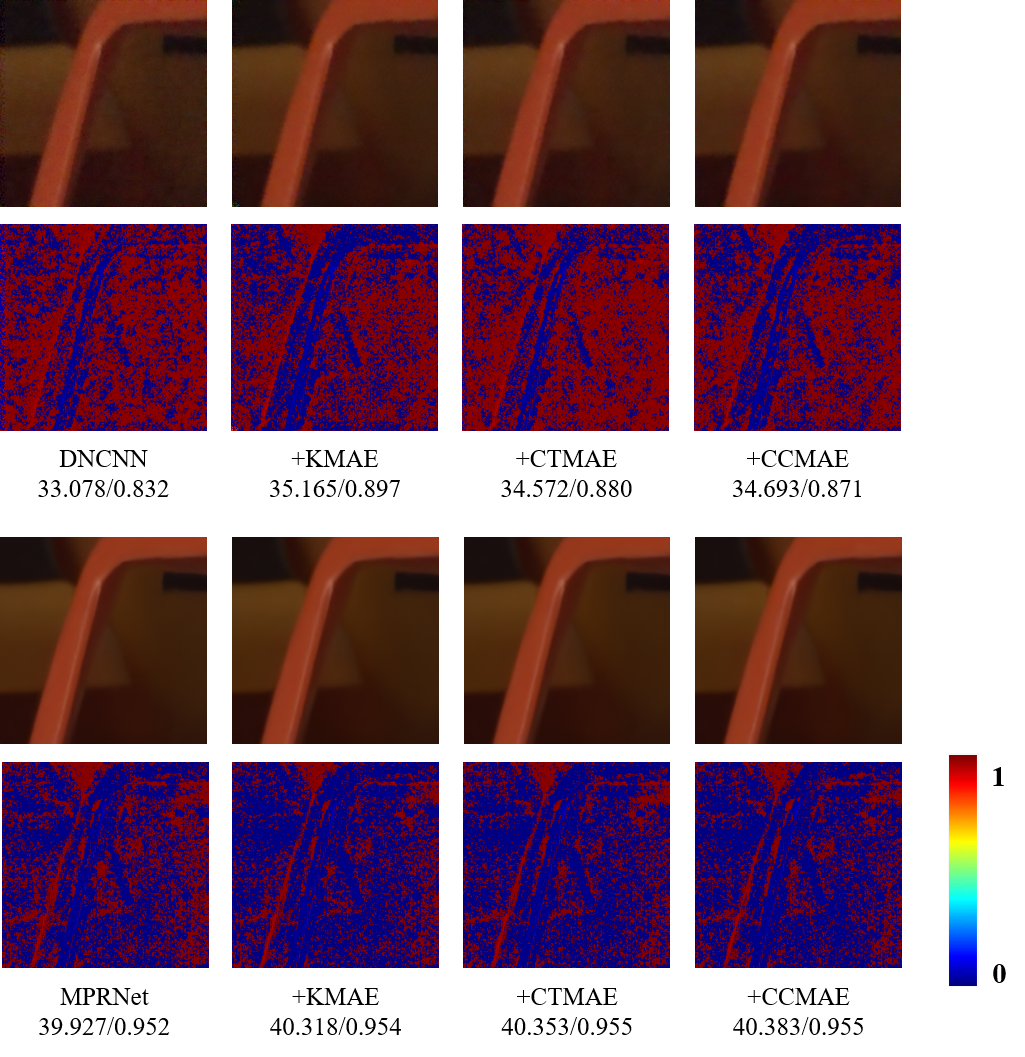}
	\caption{The visual comparison for the image de-noising. We also list the PSNR/SSIM scores under each case.}
	\label{dn2}
\end{figure}

\begin{figure}[t]
	\centering
	\includegraphics[width=\columnwidth]{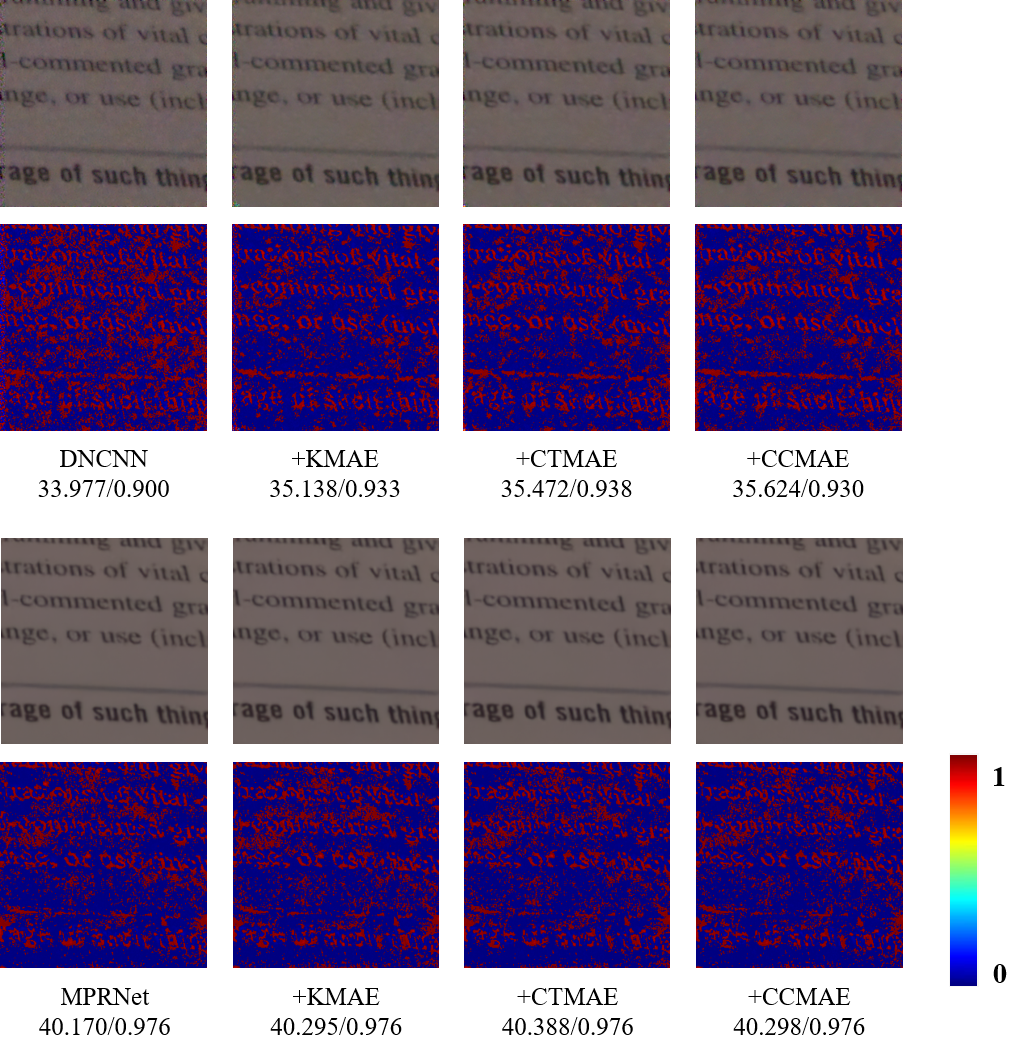}
	\caption{The visual comparison for the image de-noising. We also list the PSNR/SSIM scores under each case.}
	\label{dn3}
\end{figure}

\begin{figure}[t]
	\centering
	\includegraphics[width=\columnwidth]{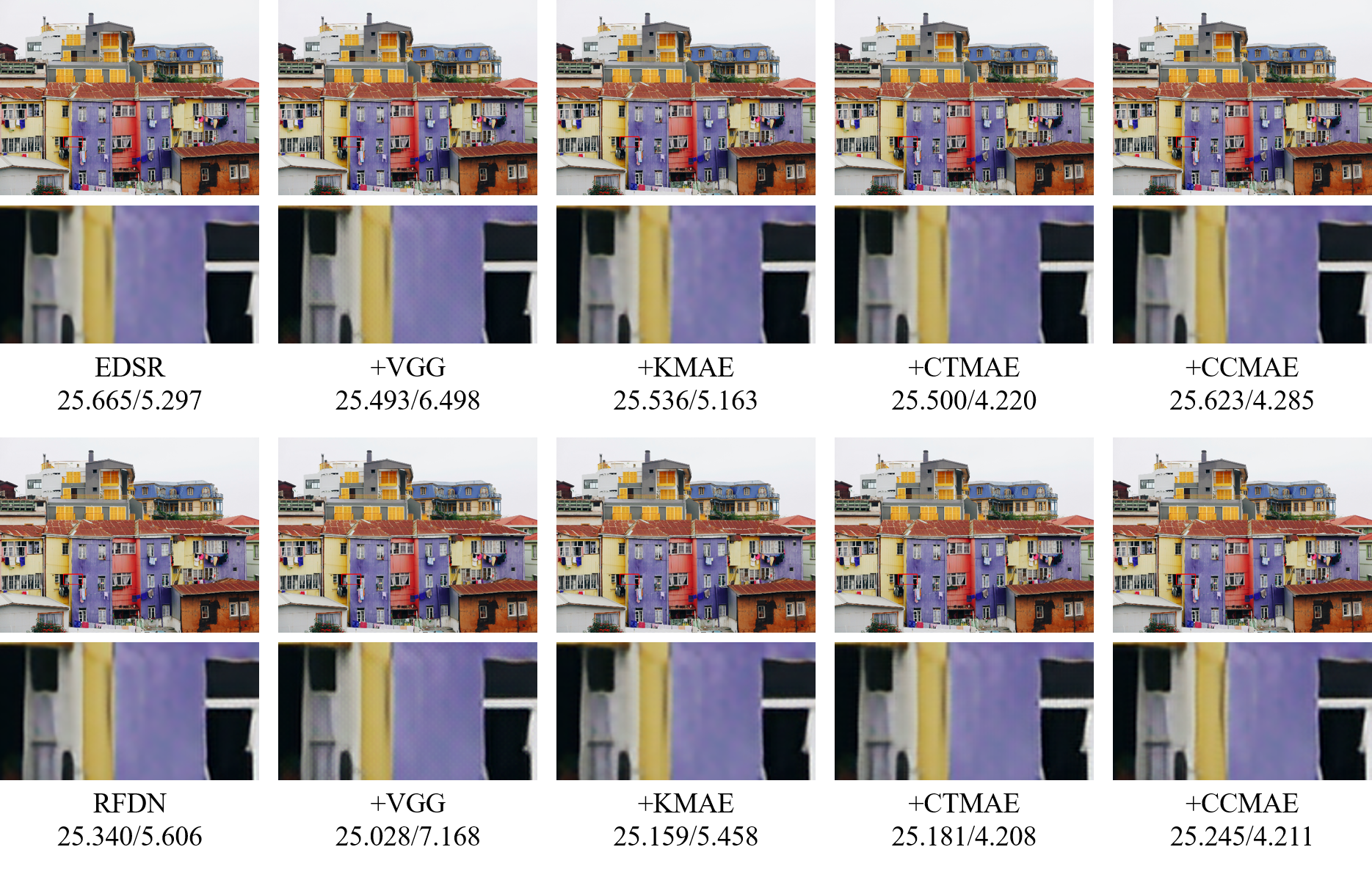}
	\caption{The visual comparison for the image super-resolution. We also list the PSNR/NIQE scores under each case.}
	\label{sr1}
\end{figure}

\begin{figure}[t]
	\centering
	\includegraphics[width=\columnwidth]{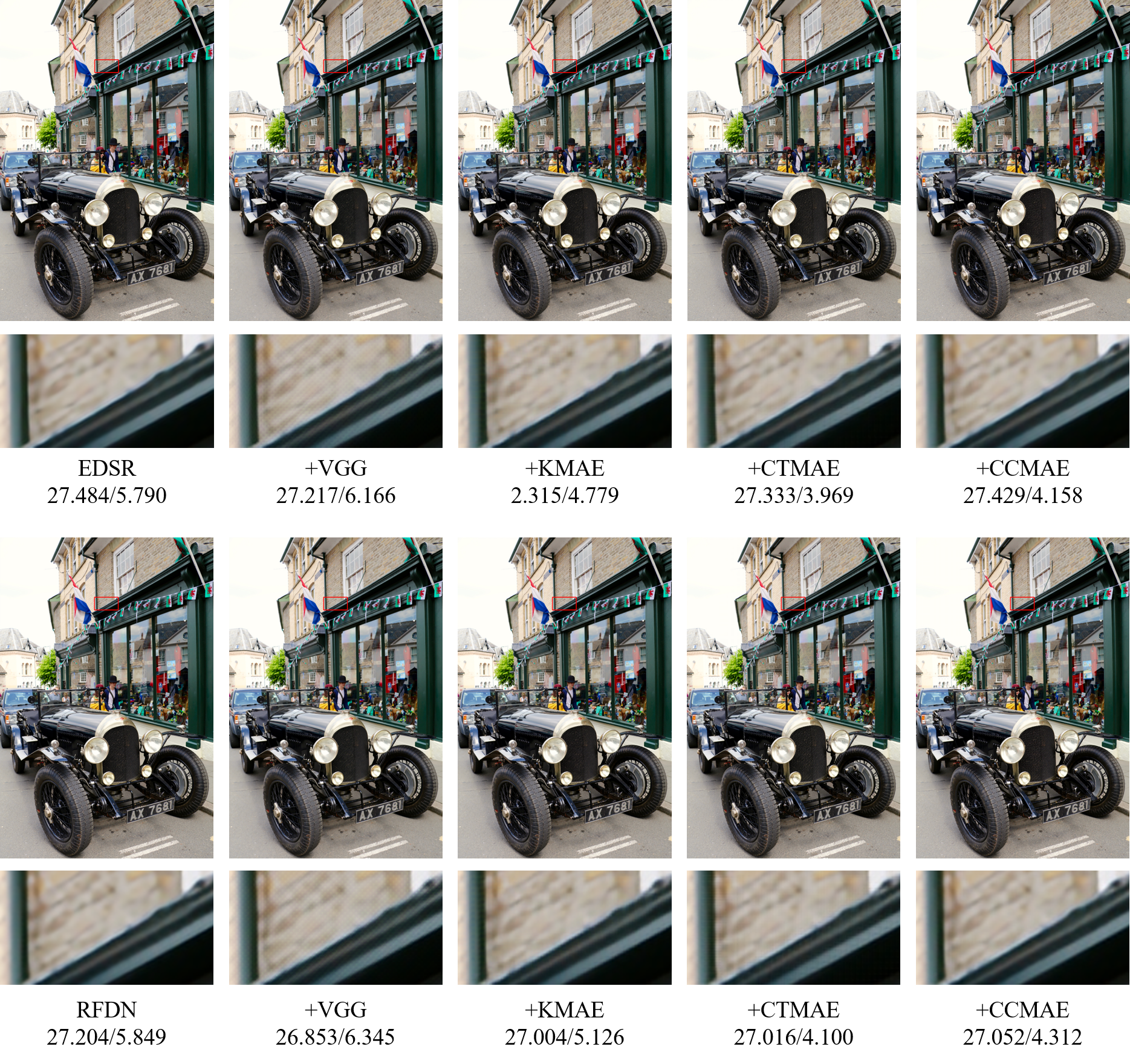}
	\caption{The visual comparison for the image super-resolution. We also list the PSNR/NIQE scores under each case.}
	\label{sr2}
\end{figure}

\begin{figure}[t]
	\centering
	\includegraphics[width=\columnwidth]{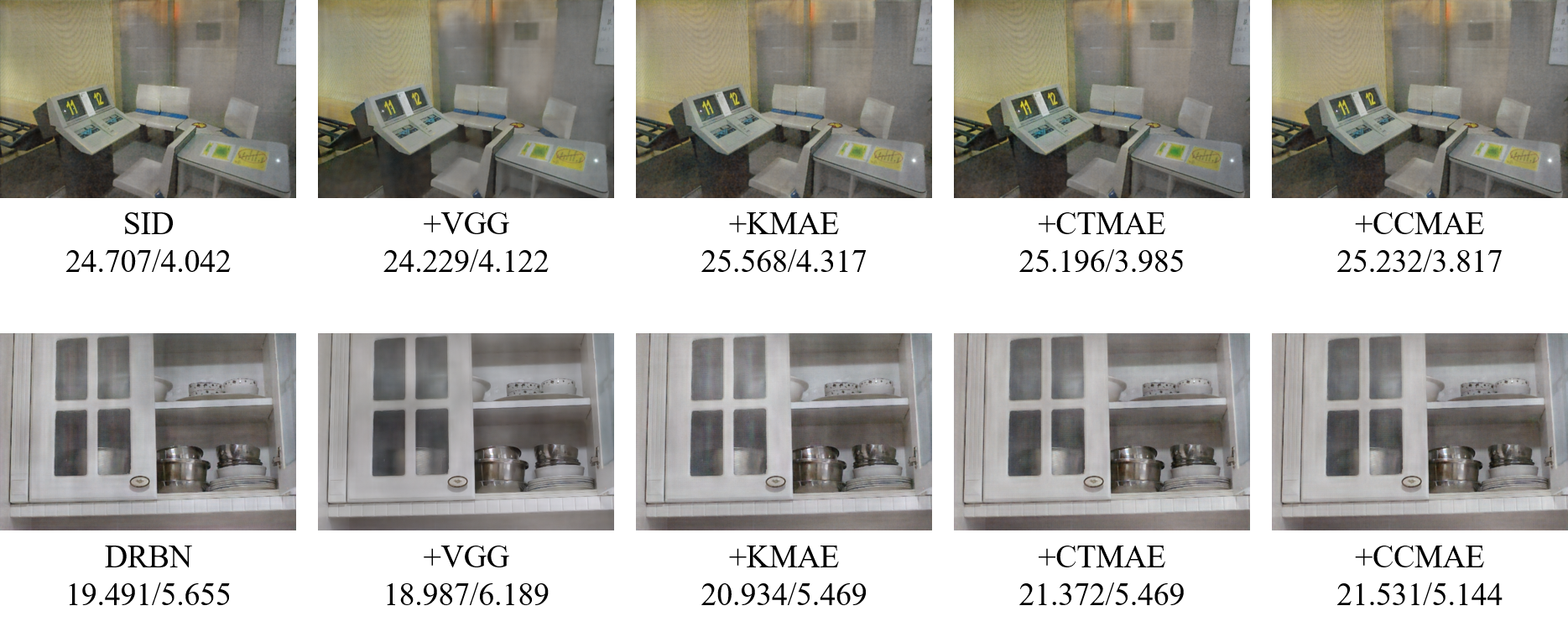}
	\caption{The visual comparison for the low-light image enhancement. We also list the PSNR/NIQE scores under each case.}
	\label{le}
\end{figure}

\begin{figure}[t]
	\centering
	\includegraphics[width=\columnwidth]{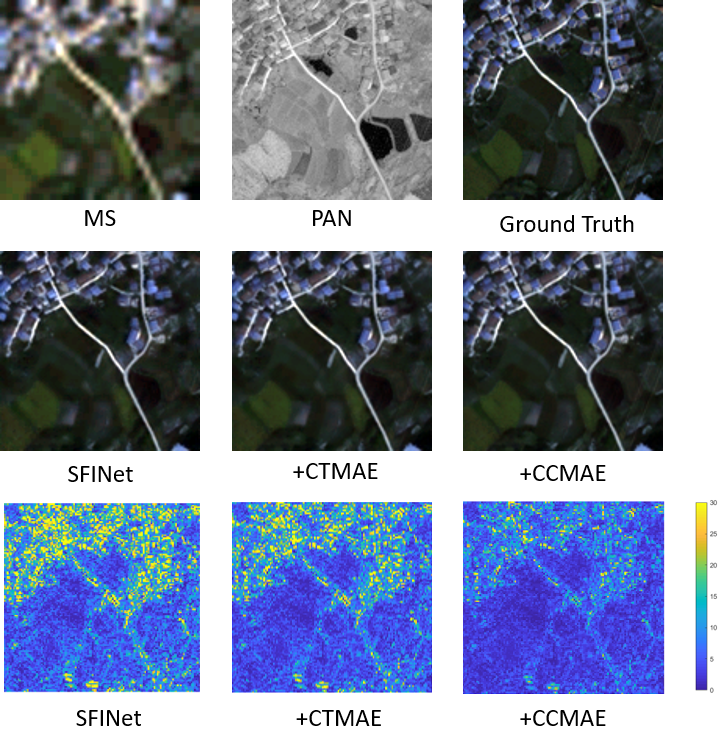}
	\caption{The visual comparison for the pan-sharpening.  Images in the last row visualizes the MSE between the pan-sharpened results and the ground truth. (Please
zoom in to see more details.}
	\label{ps}
\end{figure}

Due to the page limits, the submitted manuscript has not presented the sufficient visual results of the reported tasks over the reported baselines. In the supplementary materials, we provide the representative samples to validate the effectiveness of our belief over image de-noising task of Figure \ref{dn1}, Figure \ref{dn2} and Figure \ref{dn3},  image super-resolution of Figure \ref{sr1} and Figure \ref{sr2}, low-light image enhancement of Figure \ref{le}, pan-sharpening of Figure \ref{ps}. As can be seen, integrating with our belief is capable of improving the visual quality.

\begin{figure*}[t]
	\centering
	\includegraphics[width=\textwidth]{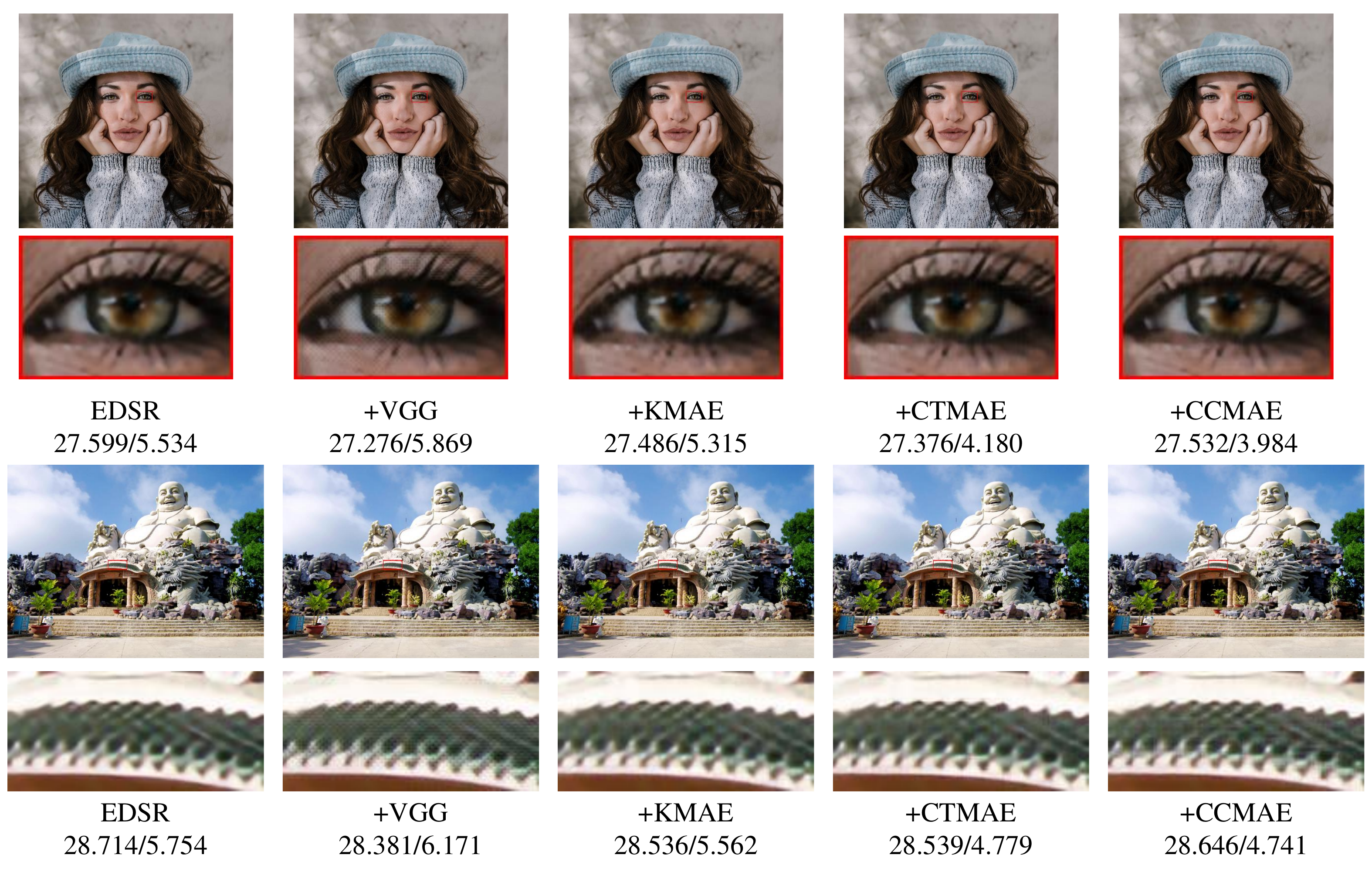}
	\caption{The visual comparison on $\times 4$ times image super-resolution. We also show the PSNR/NIQE scores under each case.  Zoom in for best view.}
	\label{fig:graphagg}

\end{figure*}

\section{Perceptual-distortion trade-off of VGG loss.}
\label{noisePrior}
\begin{figure}[t]
	\centering
	\includegraphics[width=\columnwidth]{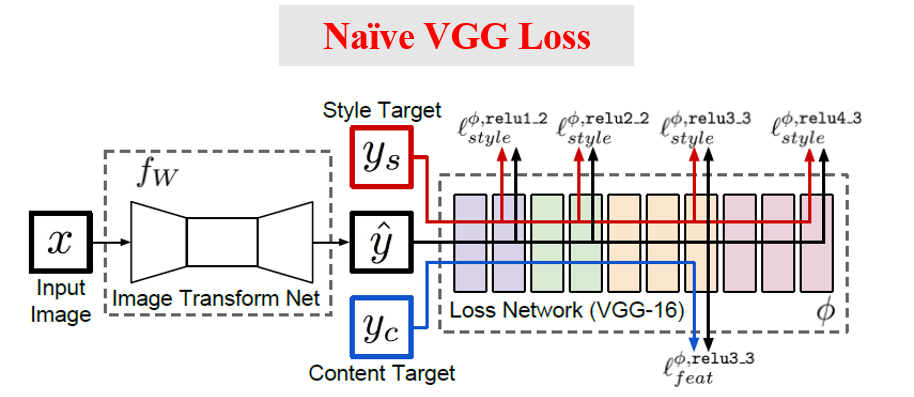}
	\caption{The implementation details of VGG loss.}
	\label{vggloss}
\end{figure}

\begin{figure}[t]
	\centering
	\includegraphics[width=\columnwidth]{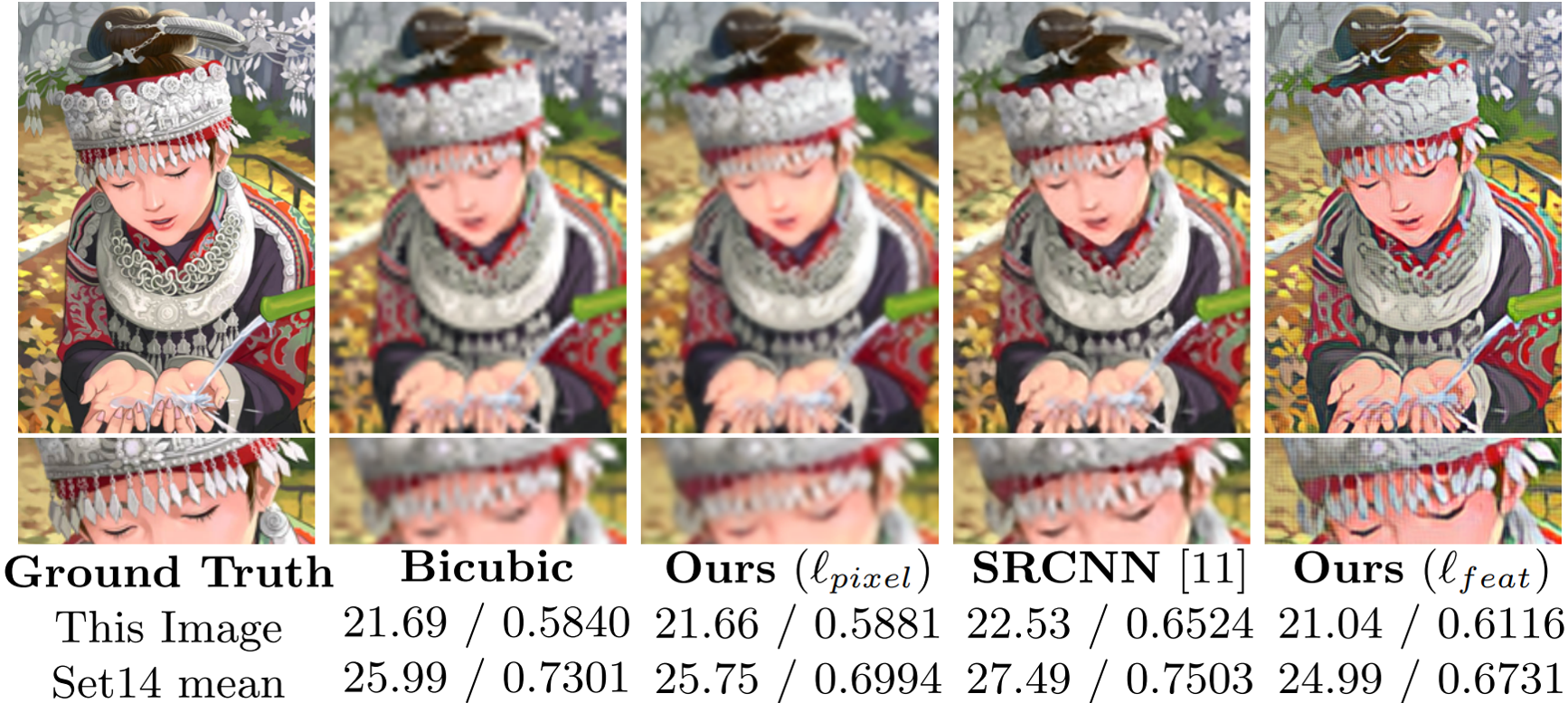}
	\caption{The perceptual-distortion trade-off of VGG loss.}
	\label{vgg}
\end{figure}

\begin{figure}[t]
	\centering
	\includegraphics[width=\columnwidth]{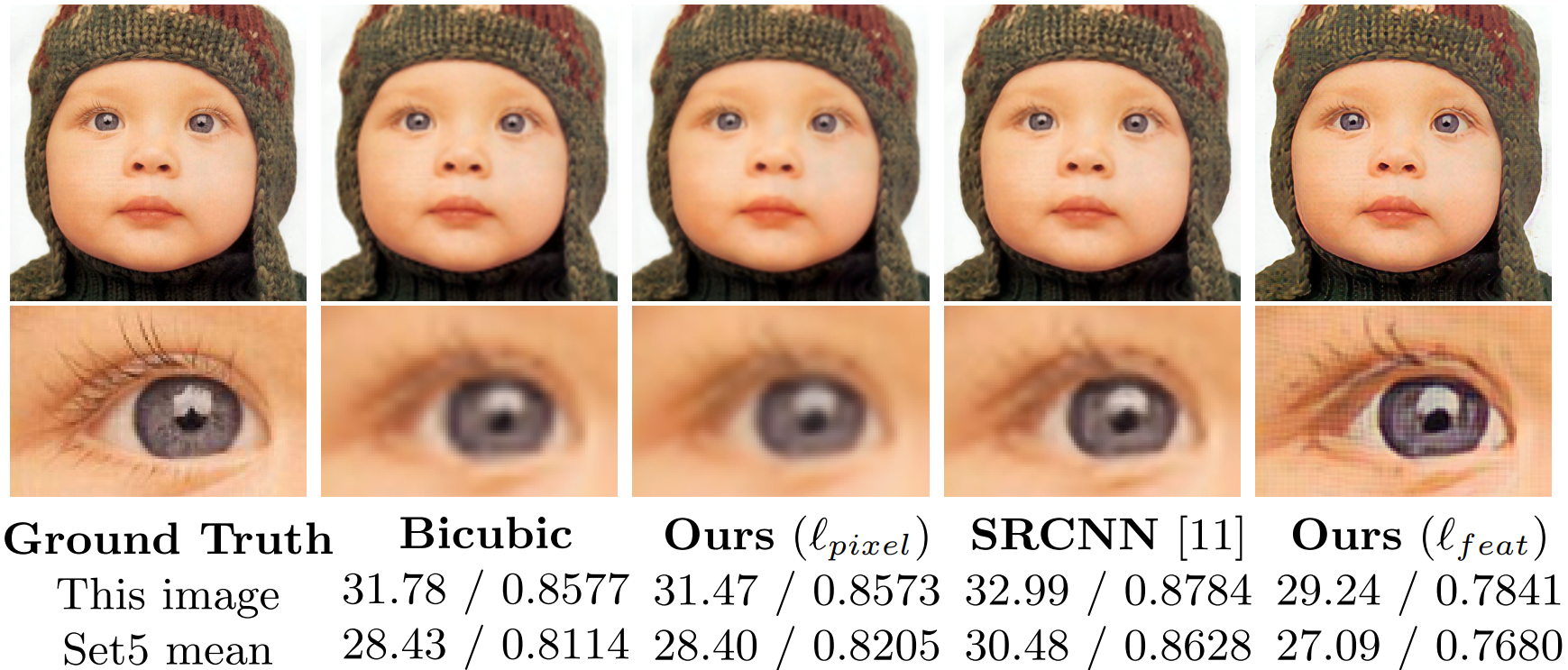}
	\caption{The perceptual-distortion trade-off of VGG loss.}
	\label{vgg1}
\end{figure}

As shown in Figure \ref{vggloss}, perceptual loss  employs the VGG network pre-trained on ImageNet as the perceptual measurement function. Since the VGG network can represent perceptual and semantic information via pre-training on the large-scale dataset, it intuitively has the capability of transferring such knowledge to downstream tasks when it is used as the loss function. However, the perceptual loss suffers from the perceptual-distortion trade-off, i.e., improving the visual performance at the cost of the PSNR and SSIM scores. The corresponding visual clues are presented in Figure \ref{vgg} and Figure \ref{vgg1}  where the PSNR and SSIM also are reported. To emphaseize, the VGG loss results in the negative grid effect.

\section{Patch-version designs.}
\label{ablation}

In addition to feeding the whole model output and ground
truth into the pre-trained MAE for measurement, the alternative strategy is to randomly select the image patch of model output and ground truth and feed them into the pre-trained MAE. It can be treated as “regressive-version patch generative adversarial networks”.

\begin{figure}[t]
	\centering
	\includegraphics[width=\columnwidth]{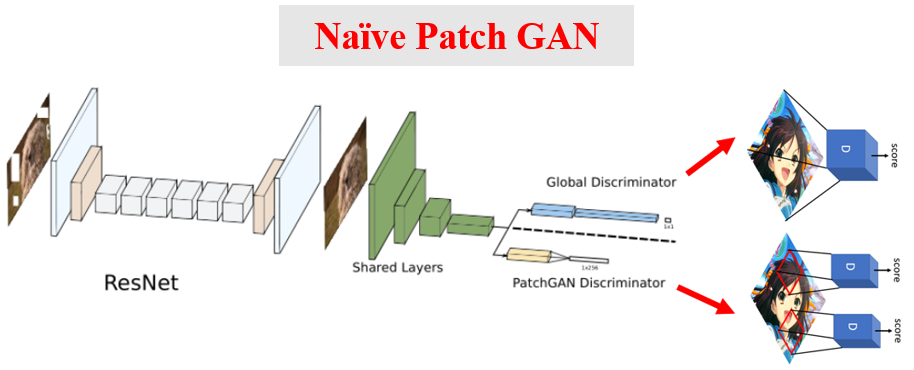}
	\caption{The implementation details of patch GAN.}
	\label{patch}
\end{figure}

As shown in Figure \ref{patch}, the original patch generative adversarial networks receive the image patch generated from the model output and ground truth and then employ the classification loss to distinguish the real and fake objective. However, the image belongs to the Null set. Directly employing the “classification-version patch generative adversarial networks”  is not suitable for image restoration task. To this end, patch-version MAE designs can be treated as “regressive-version patch generative adversarial networks” and thus benefit the image restoration tasks.

\section{Broader impacts.}

Our work shows the promising capability of the learned loss function for computer vision algorithms to empower the performance gains. Integrating our belief of learned loss function will improve the performance of neural networks and facilitate  the development of AI in real-world applications. However, the efficacy of our method may raise potential concerns when it is improperly used. For example, the safety of the applications of our  method in real-world applications may not be guaranteed. We will investigate the robustness and  effectiveness of our method in broader real-world applications.

\section{Limitations} 
First, the more comprehensive experiments on broader computer vision  tasks (\emph{e.g.}, image de-blurring) have not been explored. Second, more experiments on representative baselines are missed and need to be conducted.  Note that, orthogonal to the existing data and model studies, the focus of this work is beyond proposing a novel loss function paradigm to empower the model learning capability, sparking the realms of loss function.

\section{Broader impacts.}

Our work shows the promising capability of the learned loss function for computer vision algorithms to empower the performance gains. Integrating our belief of learned loss function will improve the performance of neural networks and facilitate  the development of AI in real-world applications. However, the efficacy of our method may raise potential concerns when it is improperly used. For example, the safety of the applications of our  method in real-world applications may not be guaranteed. We will investigate the robustness and  effectiveness of our method in broader real-world applications.

\end{document}